\documentclass[journal]{IEEEtran}
\usepackage[nolist]{acronym}
\usepackage{chemmacros}
\usepackage{multirow}
\usepackage{tikz}
\usepackage{xcolor}
\usepackage{pgfplots}
\pgfplotsset{compat=1.18}
\usepackage{graphicx}

\usepackage[group-separator={,},group-minimum-digits=4]{siunitx}
\DeclareSIUnit\year{y}
\DeclareSIUnit\pts{points}

\definecolor{Unchanged}{RGB}{67,1,84}
\definecolor{NewBuild}{RGB}{0,183,255}
\definecolor{Demol}{RGB}{0,12,235}
\definecolor{VegeN}{RGB}{0,217,33}
\definecolor{VegeG}{RGB}{255,230,0}
\definecolor{VegeR}{RGB}{255,140,0}
\definecolor{MOch}{RGB}{255,0,0}

\definecolor{Ground}{RGB}{0,21,181}
\definecolor{Building}{RGB}{51,139,255}
\definecolor{Vegetation}{RGB}{84,219,0}
\definecolor{MO}{RGB}{255,196,0}
\usepackage{cite}

\ifCLASSINFOpdf
\else
\fi
\usepackage{url}

\begin{document}
%
\title{Change detection needs change information: improving deep 3D point cloud change detection}
%
%
%

\author{Iris de~G\'{e}lis, Thomas Corpetti, S\'{e}bastien Lef\`{e}vre

\thanks{Iris de~G\'{e}lis and S\'{e}bastien Lef\`{e}vre are with the Institut de Recherche en Informatique et Syst\`{e}mes Al\'{e}atoires (IRISA), UMR 6074, Universit\'{e} Bretagne Sud, Vannes, France (e-mail: iris.de-gelis@irisa.fr,sebastien.lefevre@irisa.fr). Iris de~G\'{e}lis is also with Magellium, Toulouse, France. Her work is partly funded by the CNES, Toulouse, France. Thomas Corpetti is with the Littoral -- Environnement -- T\'{e}l\'{e}d\'{e}tection -- G\'{e}omatique (LETG), UMR 6554, Universit\'{e} Rennes 2, Rennes, France. (e-mail: thomas.corpetti@cnrs.fr).}

}

\maketitle

\begin{abstract}
Change detection is an important task that rapidly identifies modified areas, particularly when multi-temporal data are concerned. In landscapes with a complex geometry (e.g., urban environment), vertical information is a very useful source of knowledge that highlights changes and classifies them into different categories. In this study, we focus on change segmentation using raw three-dimensional (3D) point clouds (PCs) directly to avoid any information loss due to the rasterization processes. While deep learning has recently proven its effectiveness for this particular task by encoding the information through Siamese networks, we investigate herein the idea of also using change information in the early steps of deep networks. To do this, we first propose to provide a Siamese KPConv state-of-the-art (SoTA) network with hand-crafted features, especially a change-related one, which improves the mean of the Intersection over Union (IoU) over the classes of change by 4.70\%. Considering that a major improvement is obtained due to the change-related feature, we then propose three new architectures to address 3D PC change segmentation: OneConvFusion, Triplet KPConv, and Encoder Fusion SiamKPConv. All these networks consider the change information in the early steps and outperform the SoTA methods. In particular, Encoder Fusion SiamKPConv overtakes the SoTA approaches by more than 5\% of the mean of the IoU over the classes of change, emphasizing the value of having the network focus on change information for the change detection task. The code is available at \url{https://github.com/IdeGelis/torch-points3d-SiamKPConvVariants}.
\end{abstract}


%
\IEEEpeerreviewmaketitle

\section{Introduction}
\IEEEPARstart{I}{n} an ever-evolving world, being able to sense landscape transformations is of prime importance. The change detection task aims to highlight these modifications from two or more successive observations. Its application in the urban or geoscience domains are numerous, e.g., to easily update maps \cite{champion20102d}, identify damaged areas in case of natural disasters \cite{sublime2019automatic,zahs2023classification}, help city managers \cite{sandric2007change,feranec2007corine}, highlight coastal modifications \cite{letortu_retreat_2015,enriquez2019assessing, degelis2022cliff}, identify glacier melting \cite{hock2005glacier}, and detect landslides \cite{malamud2004landslide}.

Whether for urban \cite{stilla2023change,xiao20233d} or geoscience \cite{qin20163d} application, \ac{3D} data, such as \acp{PC} appear interesting because they provide additional vertical information not available in \ac{2D} images, thereby allowing a better characterization of the complex landscape geometry. In practice, a \ac{PC} is an unordered and sparse set of points represented by their \ac{3D} coordinates in a frame of reference (e.g., Cartesian coordinate system). For Earth observation purposes, \ac{3D} \acp{PC} acquired via the photogrammetric process or through \ac{LiDAR} sensors (e.g., through \ac{ALS} for example) are becoming widespread. The specific characteristics of \acp{PC} often lead to rasterization into a 2.5D \ac{DSM} that can be easily handled using traditional image processing methods \cite{murakami1999change,stal2013airborne,zhang2018change,zhang2019detecting}. However, rasterization implies significant information loss, which can be prejudicial; thus, an increasing number of studies is encouraging the design of methods that can deal with raw \ac{3D} \acp{PC} \cite{bernard2021beyond,degelis2023siamese}.

We recently \cite{degelis2023siamese} showed the possibilities brought by deep learning networks in terms of performing change detection and characterization based on raw \acp{PC}. Built upon \ac{2D} image change detection deep learning methods \cite{daudt2018fully} and \ac{3D} \acp{PC} convolutions \cite{thomas2019kpconv}, the Siamese \ac{KPConv} network outperforms the traditional distance \cite{girardeau2005change, lague2013accurate}, \ac{DSM} \cite{zhang2018change,zhang2019detecting} and machine learning-based methods \cite{tran2018integrated} on both real and simulated datasets for multiple change segmentation in urban areas. Aside from this contribution, the literature remains sparse in deep learning methods tackling the change detection at point level, i.e., segmentation of each point of one \ac{PC} according to the different type of changes compared to the other \ac{PC}. One can cite the work of \cite{zhang2018change,zhang2019detecting} that applied deep models to retrieve changes; however, they only focused on 2.5D \ac{DSM}. By contrast, \cite{KU2021192} proposed the processing of a raw \ac{3D} \ac{PC}, thanks to graph convolutions \cite{wang2019dynamic}. However, their method was designed for the change classification task, that is, retrieving changes at the scene level, as proposed by the Change3D \cite{KU2021192} or \ac{Urb3DCD-Cls} \cite{degelis2023siamese} datasets. This task is less precise than the change segmentation task because it allows identifying only the main changed object in a scene without precisely localizing it. Hence, the design of new methods that can enhance change segmentation in raw \ac{3D} \acp{PC} opens up new perspectives.

Considering the limited literature of deep learning methods for \ac{3D} \ac{PC} change detection, all three studies \cite{degelis2023siamese, zhang2019detecting, KU2021192} relied on Siamese architecture given its ability to detect changes in \ac{2D} remote sensing \cite{zhan2017change,lefevre2017toward, he2018matching, daudt2018fully,shi2020change}. Recent studies on \ac{2D} change detection also showed that data fusion is a crucial step in change detection. Therefore, paying more attention on how to fuse information coming from two network inputs and how to incorporate this fused information (i.e., change information) can improve the change detection results. Multi-temporal fusion leads to better results when performed at multiple scales \cite{daudt2018fully, chen2019deepSiam, zhang2020feature, zheng2023global}. While \cite{daudt2018fully} proposed the merging of information from both branches either by feature concatenation or differentiation, others proposed more advanced fusion modules. For example, \cite{song2021suacdnet} presented a network based on the three results of feature addition, subtraction, and concatenation at multiple scales.  \cite{jiang2022joint} took a step aside from the traditional Siamese network with one input for each branch by proposing to take the image concatenation in one branch and the difference in the other to form two sub-networks with different properties. The features of the two branches were summed at the output of each layer and concatenated at the corresponding scale in the decoder, thanks to skip connections. \cite{YIN2023103206} proposed to embed different fusion modules relying both on the multi-scale feature difference aggregation and attention on the bi-temporal feature concatenation. According to \cite{peng2020optical}, considering both the concatenation and difference of input images is more efficient even in single-stream methods. As with multi-scale, another method category uses attention mechanism to help the network focus on the most important features for the multi-temporal information fusion \cite{jiang2020pga, chen2021attention, song2021suacdnet, chen2022msf}. For \ac{3D} \ac{PC}, how to fuse the data from both epochs is not immediately clear because no point-to-point matching exists between the two \acp{PC} even in unchanged areas. To deal with this issue, Siamese \ac{KPConv} includes a nearest-point difference of encoder features at multiple scales. These feature differences are directly integrated into the decoder. Thereby, encoders do not process change information but only mono-date \ac{PC} without any information on the other \ac{PC}.  We believe that processing change information earlier in the network will boost deep networks toward the final change detection and categorization at point level.

We propose herein the enhancement of the \ac{3D} \ac{PC} change segmentation results based on the consideration that particular attention should be given to change information (i.e., the fusion of information from both dates). We perform the following steps to achieve this enhancement: i) we perform an experiment with additional hand-crafted features, particularly a change-related one, as input along with \ac{3D} point coordinates to the existing Siamese \ac{KPConv} network; ii) we design three new deep learning architectures for the \ac{3D} change segmentation; and iii) we prove the effectiveness of the two last items on the public simulated dataset \ac{Urb3DCD} \cite{degelis2021change,degelis2023siamese} as well as on the real dataset \ac{AHN-CD} \cite{degelis2023siamese}.

The remainder of this paper is structured as follows: Section~\ref{met} describes the hand-crafted features and the three new architectures; Section~\ref{res} provides a method assessment; Section~\ref{discu} discusses the obtained experimental results; and Section~\ref{sec:ccl} concludes this work.

\section{Incorporating change information in deep models}\label{met}

To incorporate the change information in deep networks, we first propose to add some hand-crafted features, also called engineered features, and one change-related feature (Sec.~\ref{sec:hf}) as input to the current state-of-the-art methods for the \ac{3D} \ac{PC} change detection. We then propose three new deep architectures (Sec.~\ref{sec:models}) directly integrating the change information into the encoder and conversely to Siamese \ac{KPConv} \cite{degelis2023siamese}.

\subsection{Considering hand-crafted features}\label{sec:hf}

Even if this is not usual in deep learning, some studies showed that combining deep and hand-crafted features improves the final results in computer vision \cite{nanni2017handcrafted} and in remote sensing \cite{nijhawan2019hybrid}. \cite{hsu2020incorporating} showed that incorporating hand-crafted features into a deep learning framework allows the improvement of the \acp{PC} semantic segmentation. They evaluated the benefit of giving different feature types in addition to \ac{3D} point coordinates for PointNet \cite{qi2017pointnet} and PointNet++ \cite{qi2017pointnet++} \ac{3D} deep frameworks. Depending on the dataset (i.e., \ac{MLS} or \ac{ALS}), the PointNet basic architecture can match or even outperform more complex architectures, such as PointNet++ and \ac{KP-FCNN} \cite{thomas2019kpconv}, when the input embeds hand-crafted features. 

Therefore, we chose to investigate whether or not adding hand-crafted features, particularly a change-related one, in the Siamese \ac{KPConv} \cite{degelis2023siamese} deep network influences the change segmentation results.
We used the following hand-crafted features:
\begin{itemize}  
    \item[\textbullet] Point distribution represented by their organization in  their neighborhood (i.e., $L_\lambda$, $P_\lambda$, and $O_\lambda$);
    \item[\textbullet] Point normals ($N_x$, $N_y$, $N_z$);
    \item[\textbullet] Height information (i.e., $Z_{range}$, $Z_{rank}$, and $nH$);
    \item[\textbullet] Change information (i.e., $Stability$).
\end{itemize}
The information on the distribution of points contained in the neighborhood is given by the three following variables: linearity $L_\lambda$, planarity $P_\lambda$, and omnivariance $O_\lambda$. These variables represent the likelihood of a point belonging to a linear (\ac{1D}), a planar (smooth surface) (\ac{2D}) or a volumetric (\ac{3D}) neighborhood, respectively. These three attributes are common for extracting information in \ac{3D} \acp{PC}. They are computed from the three eigenvalues ($\lambda_1 \geq\lambda_2 \geq \lambda_3\geq 0$) obtained after applying a \ac{PCA} to a matrix containing the \ac{3D} coordinates of points in the neighborhood. $L_\lambda$,  $P_\lambda$ and  $O_\lambda$ are given in Equations~\ref{eq:L} to \ref{eq:O}, respectively.
\begin{equation}\label{eq:L}
    L_\lambda = \frac{\lambda_1 - \lambda_2}{\lambda_1}
\end{equation}
\begin{equation}\label{eq:P}
    P_\lambda =  \frac{\lambda_2 - \lambda_3}{\lambda_1}
\end{equation}
\begin{equation}\label{eq:O}
    O_\lambda = \sqrt[3]{ \lambda_1\lambda_2 \lambda_3 }
\end{equation}
In practice, if $\lambda_1$ is larger than $\lambda_2$ and $\lambda_3$, $L_\lambda$ is near 1. Only one eigenvalue is meaningful in this case. That is, only one principal axis results from the \ac{PCA}, and  the points are mainly distributed along a single axis. If $\lambda_1$ and $\lambda_2$ are larger with regard to $\lambda_3$, implying that $P_\lambda$ is near 1, the points are spread in a plane defined by the eigenvectors corresponding to $\lambda_1$ and $\lambda_2$. Lastly, $O_\lambda$ is high if each of the three eigenvalues is of equal importance. This implies that the points are scattered along the three axis in a \ac{3D} volumetric space.

Once eigenvalues ($\lambda_1$, $\lambda_2$, $\lambda_3$) are computed, point normals ($N_x$, $N_y$, $N_z$) are obtained by taking the eigenvector corresponding to the smallest eigenvalue.

$Z_{range}$ and $Z_{rank}$ provide the height information by providing the maximum height (Z coordinate) difference between the points in the neighborhood and the rank of the height of the considered points within the neighborhood. The normalized height $nH$ completes the height information by providing the difference between the height of the considered points and the local \ac{DTM} (rasterization of the \ac{PC} at the ground level).

Lastly, the $Stability$ \cite{tran2018integrated} feature provides a bi-temporal information on the considered point. It is the ratio of the number of points in the spherical neighborhood to the number of points in the vertical cylindrical neighborhood in the other PC (oriented along the vertical axis). Thus, in each point of the current \ac{PC}, $Stability$ is the ratio between the number of points in the \ac{3D} ($n_{3D}$) and \ac{2D} ($n_{2D}$) neighborhoods in the other \ac{PC}, in present : 
\begin{equation}\label{eq:stability}
    Stability = \frac{n_{3D}}{n_{2D}} \times 100
\end{equation}
Looking only at the number of points in the \ac{3D} neighborhood of each point of both \acp{PC} is enough to retrieve the changes on isolated buildings and trees. However, in dense forest areas or when different objects are close to each other, the \ac{3D} spherical neighborhood may still contain points coming from some other unchanged entity. Hence, taking the ratio with the \ac{2D} neighborhood is a way to consider the unchanged points and obtain an indicator of change and object instability. The ratio will be near 100\% if no change occurs and tends to 0\% in case of changes. We expect the $Stability$ value of vegetation to be lower. 
Most of the hand-crafted features presented in \cite{tran2018integrated} are used except for those that utilize \ac{LiDAR}'s multi-target capability, because our datasets do not contain such information (see Section~\ref{sec:dataset}).
We recall that the Siamese \ac{KPConv} architecture takes as many input features as desired, by simply modifying the number of inputs of the first layer of encoders.

\subsection{New models for 3D point cloud change detection}\label{sec:models}

We will now explore how to learn this change information through novel deep networks. We built upon the Siamese \ac{KPConv} model to propose three original architectures emphasizing the change-related features. 
The three presented architectures are based on \acf{KPConv} \cite{thomas2019kpconv} because it had been proven efficient for the change detection task in \ac{3D} \acp{PC}. We fused the features coming from both \acp{PC} by using the nearest-point difference strategy, as in Siamese \ac{KPConv}:
\begin{equation}\label{eq:npdiff}
        (\mathcal{P}_1,\mathcal{F}_1) \fminus (\mathcal{P}_2,\mathcal{F}_2)  = f_{2i} - f_{1j | j=\arg\min (\| x_{2i} - x_{1j} \|)}
\end{equation}
Thus, for \acp{PC} $\mathcal{P}_{1}$ and $\mathcal{P}_{2}$, with their corresponding features $\mathcal{F}_{1}$ and $\mathcal{F}_{2}$, the feature difference $\fminus$ was computed between features $f_{2i} \in \mathcal{F}_{2}$ of each point $x_{2i} \in \mathcal{P}_{2}$ of the second \ac{PC} and features $f_{1j} \in \mathcal{F}_{1}$ of the nearest-point $x_{1j} \in \mathcal{P}_{1}$.

\begin{figure}
    \centering
    \includegraphics[width=0.5\textwidth]{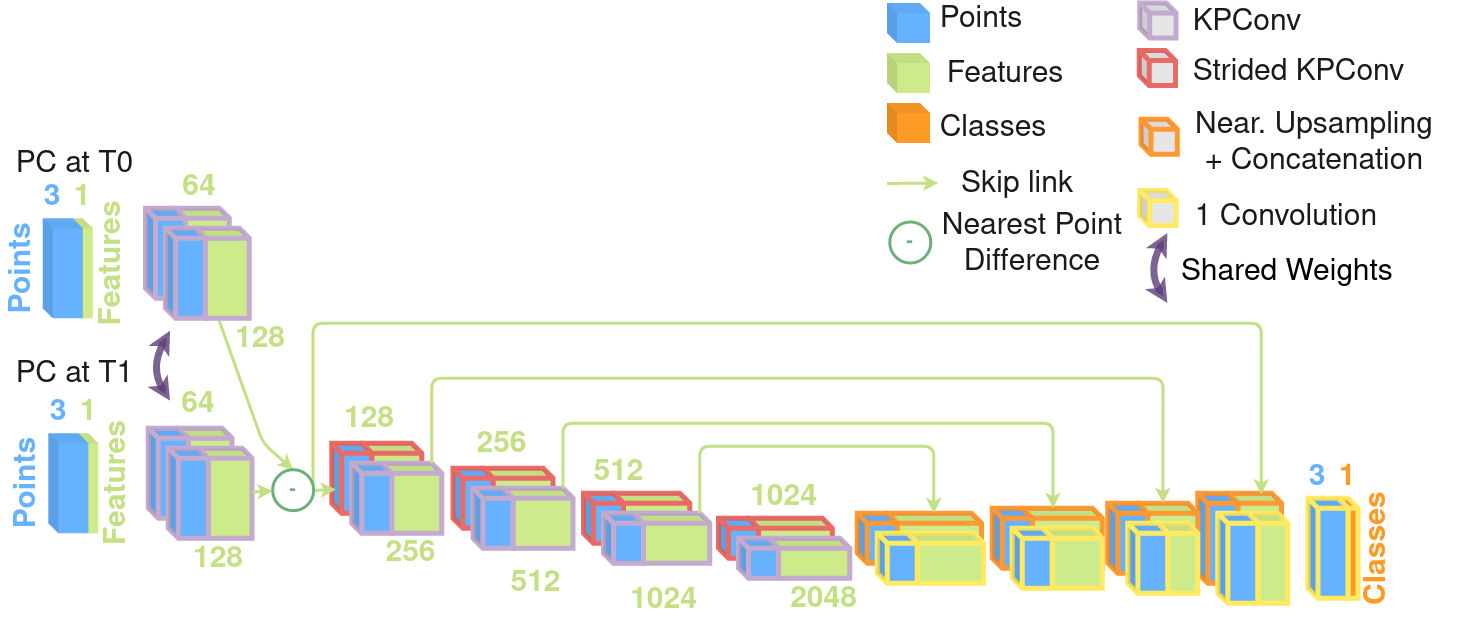}
    \caption[OneConvFusion architecture]{OneConvFusion architecture for the \ac{3D} \ac{PC} change segmentation. The links between successive layers were omitted for brevity.}
    \label{fig:archiOneConvFusion}
\end{figure}

The first option was to create a relatively simple network by fusing the information of both \acp{PC} just after the first layer (Figure~\ref{fig:archiOneConvFusion}). The following layer of the encoder took as input only the nearest-point feature difference (noted $\fminus$). The following layers of the encoder and the decoder took as input the output of the previous layer, as in a classical \ac{FCN}. Here, the idea was to evaluate the benefits of dealing with differences early in the process. This architecture is called OneConvFusion. 

\begin{figure}
    \centering
    \includegraphics[width=0.5\textwidth]{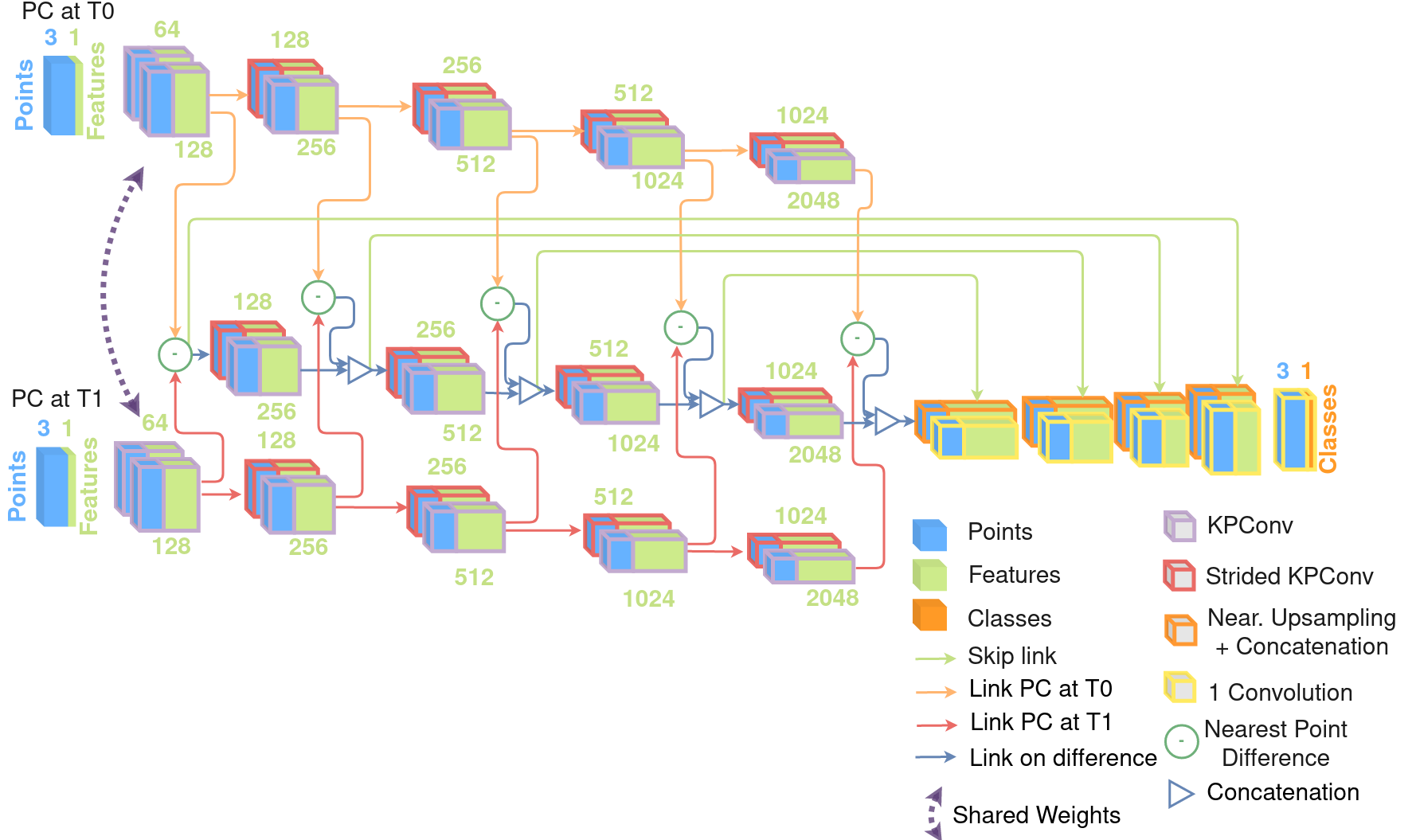}
    \caption[Triplet KPConv architecture]{Triplet KPConv architecture for \ac{3D} the \acp{PC} change segmentation.}
    \label{fig:archiTriplet}
\end{figure}

However, the mono-date features of the first layer might not be sufficient for accurate change identification. Therefore, we designed the Triplet \ac{KPConv} network (Figure~\ref{fig:archiTriplet}), that contained two encoders for extracting the mono-date information (as in the Siamese \ac{KPConv} network) and an additional encoder that extracts the change-related features. The  ``change encoder'' took as input the nearest-point difference computed after the first layer of the mono-date encoders. The following layers of the change encoder took as input the output feature concatenation of the previous layer and the result of the nearest-point features (from the mono-date encoder) difference of the corresponding scale. The multi-scale mono-date and change information were both considered. The decoder used the features extracted by the change encoder as the input. Note that mono-date encoders can either share weights or not, the latter leading to the Pseudo-Triplet \ac{KPConv}, as for Siamese \ac{KPConv}, and a Pseudo-Siamese KPConv. Therefore, in a shared-weight configuration, the Triplet \ac{KPConv} network is as symmetrical as the Siamese \ac{KPConv}.

\begin{figure}
    \centering
    \includegraphics[width=0.5\textwidth]{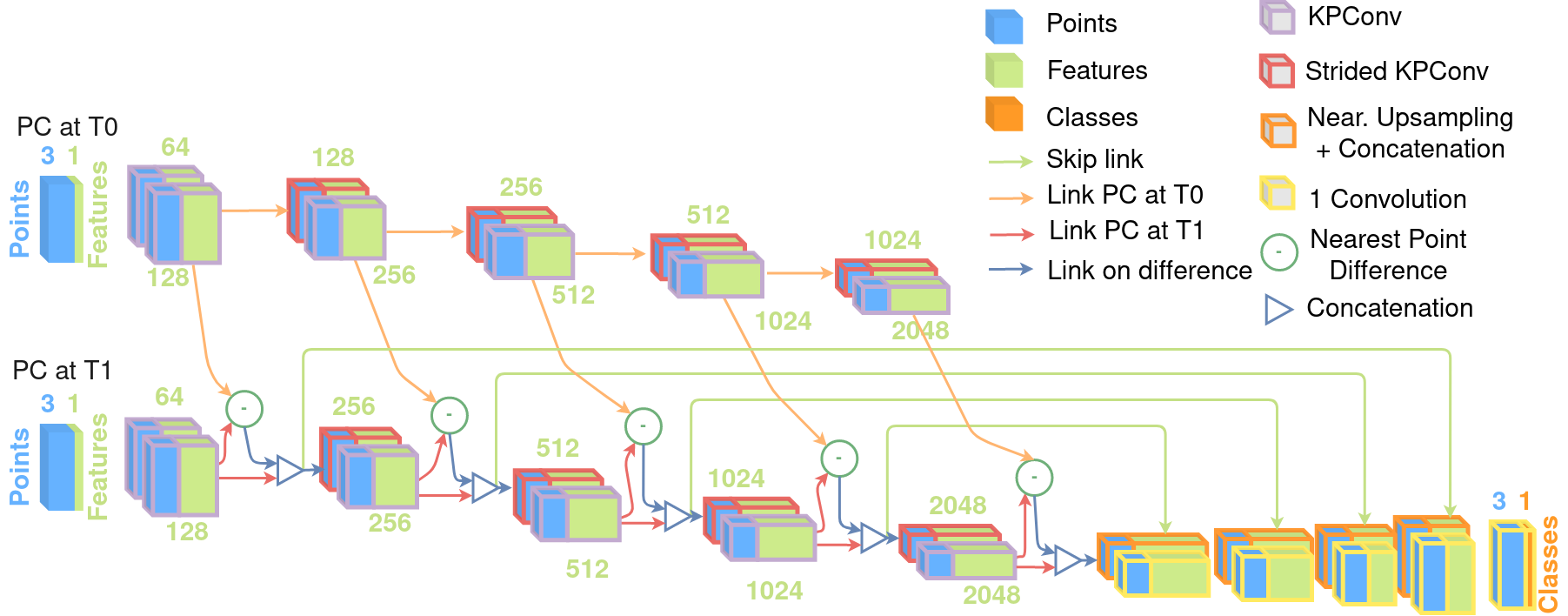}
    \caption[Encoder Fusion SiamKPConv architecture]{Encoder Fusion SiamKPConv architecture for the \ac{3D} \ac{PC} change segmentation.}
    \label{fig:archiEncFusion}
\end{figure}

The third version of the architecture was designed to directly fuse the mono-date and change features in the same encoder. This network is called Encoder Fusion SiamKPConv.  The first encoder extracted the mono-date features of the older \ac{PC} using convolution layers (top, Figure~\ref{fig:archiEncFusion}), as in all previous architectures. As illustrated in the bottom of Figure~\ref{fig:archiEncFusion}, the second encoder more specifically combined the output features from the newer \ac{PC} and the nearest point feature difference. Each layer of this second encoder took as input the output feature concatenation of the previous layer and the difference of the features from this encoder and the mono-date encoder of the older \ac{PC}. Hence, both mono-date and change features can be combined in convolutional layers. Note that this third architecture is not symmetrical, in contrast to the previous ones, because the weights of the two encoders were obviously not shared.
As regards the Triplet KPConv and OneConvFusion architectures, the idea was to encode the differences earlier in the process. In this third architecture, however, we fused them with the features of the second \ac{PC} to better combine the two different feature types.

These proposed new architectures are in line with the recent developments for the \ac{2D} image change detection in terms of the importance of data fusion \cite{chen2019deepSiam,jiang2022joint,YIN2023103206} for the change detection task. By convolving the change features in the encoder, we expect the network to put more attention on the changes and better combine the multi-scale change features.

\section{Results}\label{res}
The experimental results of our methods are elaborated here on public simulated and real datasets to quantitatively evaluate our networks. Before we describe them in detail, let us first introduce the experimental settings.

\subsection{Datasets}\label{sec:dataset}
The public dataset \acf{Urb3DCD} \cite{degelis2021change} was used for our experiments. This dataset comprised various semantic change situations inside cities based on real information related to the organization of streets, areas, etc., on which buildings, vegetation, or cars have been added. It then simulates point clouds derived from  laser pulses from \ac{ALS} with real flight plans. 

Among the different dataset versions, we chose to assess the one with a low point density (approximately \qty{.5}{\unit[per-mode=symbol]{\pts\per\m\squared}}) because it contains more change classes (i.e., `unchanged', `new building', `demolition', `new vegetation', `vegetation growth', `missing vegetation' and `mobile object') and it relies on \acp{PC} that are more realistic than the first dataset version \cite{degelis2023siamese}. 

Some quantitative results are given using the \acf{AHN-CD} dataset presented in \cite{degelis2023siamese} to prove the effectiveness of our methods on real data. This dataset is composed of two dates of the \ac{ALS} surveys over the Netherlands.  The \ac{PC} densities varied from \qtyrange[range-units = single]{10}{14}{\unit[per-mode=symbol]{\pts\per\m\squared}} for the first date (AHN3) and \qtyrange[range-units = single]{10}{24}{\unit[per-mode=symbol]{\pts\per\m\squared}} for the second date (AHN4). Some change annotations were semi-automatically derived from the mono-date semantic labeling for the training and validation sets. The test set was manually annotated to avoid any labelization errors present in the semi-automatic annotation. In addition to the `unchanged' classes, this dataset also contains the three following change classes: `new building', `demolition', and
`new clutter'.

\subsection{Experimental settings}\label{sec:siamkpconvExpeSet}

For the experimental settings, we utilized the hyperparameters used by \cite{degelis2023siamese} for Siamese \ac{KPConv}. Some cylinder pairs of \qty{50}{\m} radius were extracted from both the \acp{PC} for Urb3DCD. The first sub-sampling rate $dl_0$ was set to \qty{1}{\m}. A radius of \qty{25}{\m} was used for the \ac{AHN-CD} dataset with a first sub-sampling rate $dl_0$ set to \qty{0.5}{\m}. For training purposes, we minimized the \ac{NLL} loss using a \ac{SGD} with a 0.98 momentum. The loss was calculated as follows:
\begin{equation}\label{eq:loss}
    NLL(y_t, y_p)  = -(y_t \log(y_p) + (1-y_t) \log(1-y_p))
\end{equation}
where $y_t$ corresponds to the target label’s probability and $y_p$ is the predicted label’s probability.
Ten cylinder pairs were used in each batch. The initial learning rate was set to $10^{-2}$ and scheduled to decrease exponentially. For the training we relied on a random drawing of training cylinders as a function of the class distribution as in \cite{degelis2023siamese}, because the change detection dataset was generally imbalanced. For each training epoch, \num{6000} cylinder training pairs were seen by the network. A total of \num{3000} pairs from the validation set were used during the validation. The loss was weighted according to the class distribution to ensure learning the less-represented classes as well. Data augmentation was performed during training through the random rotation of cylinders around the vertical axis (note: both cylinders of a pair were rotated by the same angle to keep the coherence inside the pair) and the addition of a Gaussian noise at point level. 

The development of these architectures was implemented in PyTorch and relied on the \ac{KPConv} implementation available in Torch-Points3D \cite{chaton2020torch}. For the nearest-point feature difference (Equation~\ref{eq:npdiff}), the nearest point was determined by the \ac{kNN} implementation available in the \ac{GPU}-compliant PyTorch Geometric for a faster computation. 

For the hand-crafted feature extraction, the computation is made before the cylinder extraction to limit border effects. Neighborhood sizes for the $Stability$ are set at \qty{5}{\m} for Urb3DCD and \qty{3}{\m} for \ac{AHN-CD}. Concerning other neighborhoods, they are based on the 10 nearest neighbor points. Point normal and \ac{DTM} computations are performed using \ac{PDAL}\footnote{\url{https://pdal.io/en/2.5-maintenance/index.html}, accessed on 27/02/2023.}.

In change detection and categorization, datasets are largely imbalanced. In other words, most of the data belong to the unchanged class, despite this class not being the most interesting one. Hence, we preferred herein to disregard the overall accuracy or precision scores that were not very indicative of the method's performance under these settings. Accordingly, we selected the \ac{mAcc} and the mean of the \ac{IoU} over the classes of change (mIoU$_{ch}$) for a more reliable quantitative assessment of the different methods. 

All tests were conducted three times to assess the variability in the results. The average results of these three runs are given along with the standard deviation in Tables~\ref{tab:ResLid05CompaFeat}, \ref{tab:ResLid05Newarchi}, \ref{tab:ResLid05IouNewArchi} and \ref{tab:ResLid05NewarchiAHNCD}.

\subsection{Experimental results}
\subsubsection{Results of adding hand-crafted features to the Siamese KPConv network}
\begin{figure*}
    \centering
    \includegraphics[trim=3.5cm 0cm 2.9cm 0cm, clip, width=\textwidth]{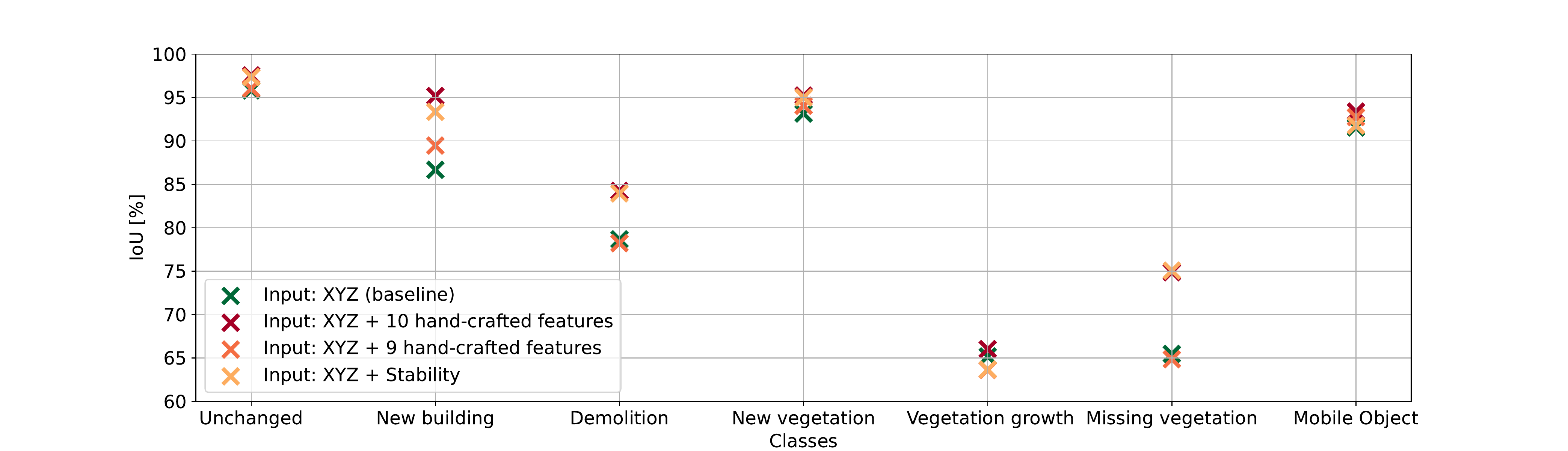}
    \caption[Influence on the IoU of adding hand-crafted features]{\textbf{Influence on per class \ac{IoU} of adding hand-crafted features along with \ac{3D} point coordinates as the input to the Siamese \ac{KPConv}.} For the `new building', `demolition' and `missing vegetation' classes, the high disparity in the \ac{IoU} demonstrated that adding the hand-crafted features as input had a larger influence compared to those on classes where the results were grouped around the same value. } 
    \label{fig:influHCfeats}
\end{figure*}

Table~\ref{tab:ResLid05CompaFeat} presents the quantitative results. Note that the results given with zero input features corresponded to those reported in the original publication of the Siamese \ac{KPConv} \cite{degelis2023siamese}. Providing the hand-crafted features as input in addition to the point coordinates considerably improved the results. We subsequently assessed the importance of the unique change-related hand-crafted feature, the $Stability$. The point distribution and the height hand-crafted features seem to have only a slight beneficial impact (+0.37\% of mIoU$_{ch}$) on the change segmentation results. In contrast, the $Stability$ feature seems to have a major impact ($+3.67$\% of mIoU$_{ch}$) on both metrics \ac{mAcc} and mIoU$_{ch}$. More specifically, when looking at the per class gain in the \ac{IoU}, the $Stability$ feature on its own principally helped in the `new building', `demolition' and `missing vegetation' classes (Figure~\ref{fig:influHCfeats}).

\begin{table}
    \centering
        \caption{\textbf{Comparison of Siamese KPConv network with different input features on the Urb3DCD-V2 low-density \ac{LiDAR} dataset.} The results are given in \%. The 10 input features are as follows: $N_x$, $N_y$, $N_z$, $L_T$, $P_T$, $O_T$, $Z_{range}$, $Z_{rank}$, $nH$ and $Stability$.}
    \label{tab:ResLid05CompaFeat}
    \footnotesize
    \begin{tabular}{cc|cc}
    \hline
         \multicolumn{2}{c}{\# of input features} & mAcc &mIoU$_{ch}$  \\ 
        \hline
         0 &&91.21 $\pm$ 0.68 & 80.12 $\pm$ 0.02 \\
          10& & \textbf{93.65} $\pm$ 0.16 & \textbf{84.82} $\pm$ 0.58 \\
          9 &w/o $Stability$ & 91.44 $\pm$ 0.47 & 80.49 $\pm$ 0.64 \\
          1 &$Stability$ only& 92.92 $\pm$ 0.47  & 83.80 $\pm$ 0.89  \\
        \hline
    \end{tabular}
\end{table}

\subsubsection{Results of the Siamese KPConv evolution}
\begin{figure*}
    \centering
    \includegraphics[trim=2.5cm 0cm 2.9cm 0cm, clip, width=\textwidth]{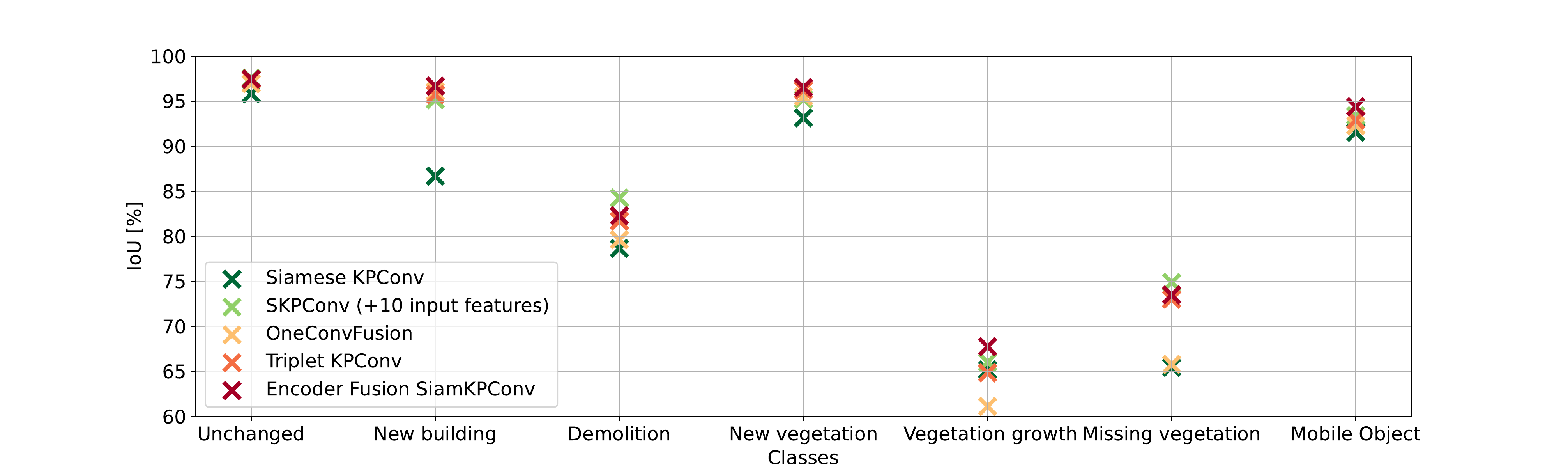}
    \caption[Influence on the IoU of the three Siamese KPConv evolutions]{\textbf{Influence on per class \ac{IoU} of the three Siamese \ac{KPConv} evolutions}, namely OneConvFusion, Triplet \ac{KPConv} and Encoder Fusion SiamKPConv. The results of Siamese \ac{KPConv} with 10 hand-crafted input features were also included for comparison purposes. } 
    \label{fig:influArchi}
\end{figure*}

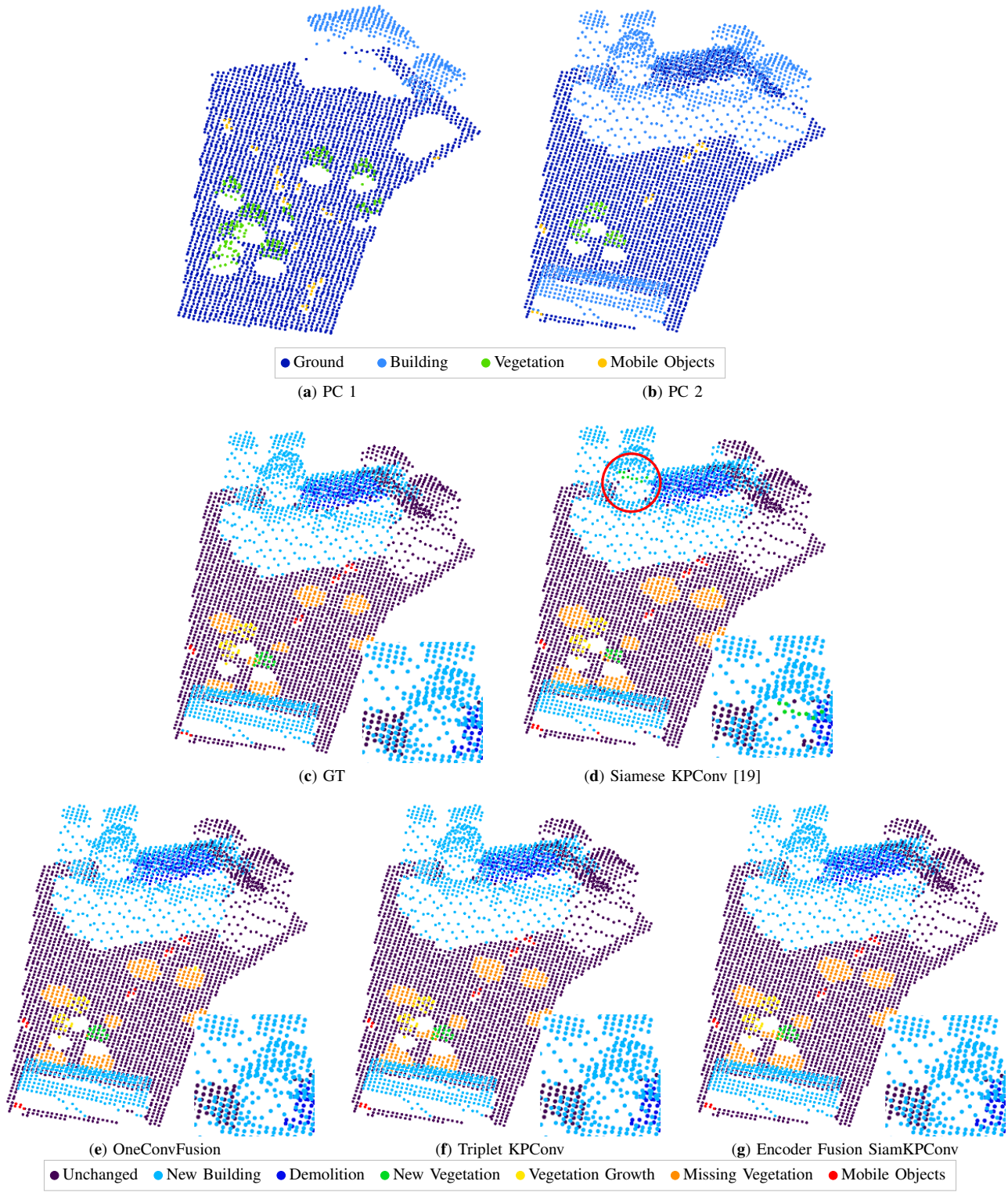
\begin{figure*}
    \centering
    \footnotesize
    \begin{tabular}{cc}
        \includegraphics[width=0.3\textwidth]{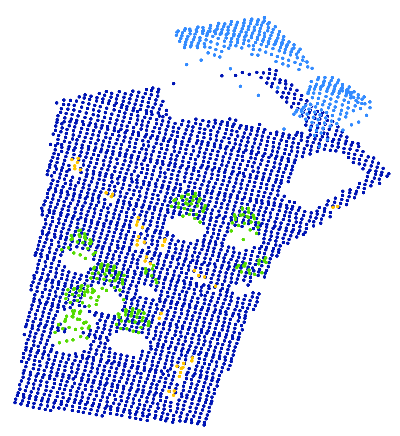} & 
        \includegraphics[width=0.3\textwidth]{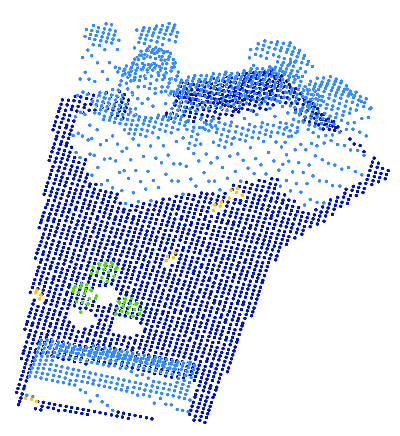} \\
        \multicolumn{2}{c}{        \begin{tikzpicture}
    		\begin{axis}[
            		xmin=1,
                    xmax=1,
                    ymin=1,
                    ymax=1,
                     hide axis,
    				width=0.9\textwidth ,
    				mark=circle,
    				scatter,
    				only marks,
    				legend cell align={left},
    				legend entries={Ground, Building, Vegetation, Mobile Objects},
    				legend style={draw=lightgray,at={(0,0)}, legend columns=4,/tikz/every even column/.append style={column sep=0.5cm}}]
    			\addplot[Ground] coordinates {(0,0)}; 
    			\addplot[Building] coordinates {(0,0)};
    			\addplot[Vegetation] coordinates {(0,0)};
    			\addplot[MO] coordinates {(0,0)};
    		\end{axis}
	    \end{tikzpicture}} \\
        (\textbf{a}) PC 1 & (\textbf{b}) PC 2 \\
        \end{tabular}
        \begin{tabular}{cc}
                \includegraphics[width=0.3\textwidth]{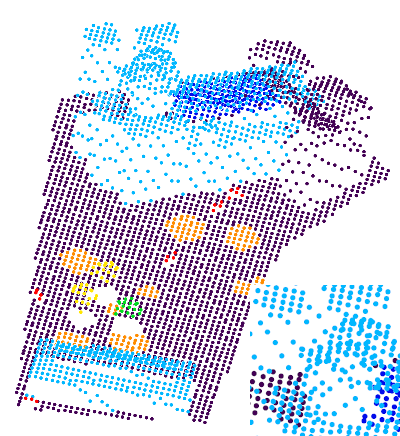}&
        \begin{tikzpicture}
    \node (image) at (0,0) {\includegraphics[width=0.3\textwidth]{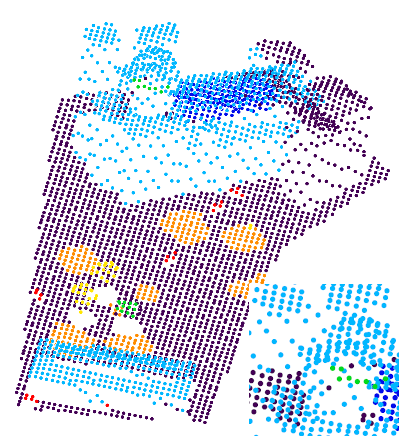}}; 
       \draw[red, very thick,rotate=0] (-0.7,1.7) ellipse (14pt and 14pt);
        
 \end{tikzpicture}\\
         (\textbf{c}) GT & (\textbf{d}) Siamese KPConv \cite{degelis2023siamese} \\
        \end{tabular}
         \begin{tabular}{ccc}
                     \includegraphics[width=0.3\textwidth]{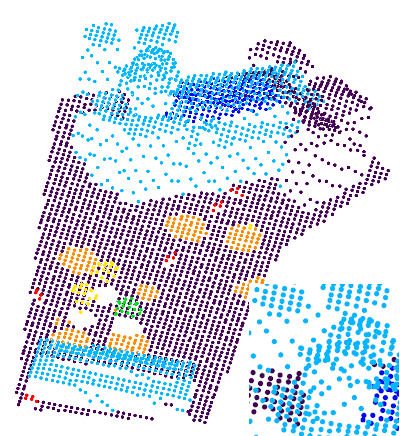}& 
        \includegraphics[width=0.3\textwidth]{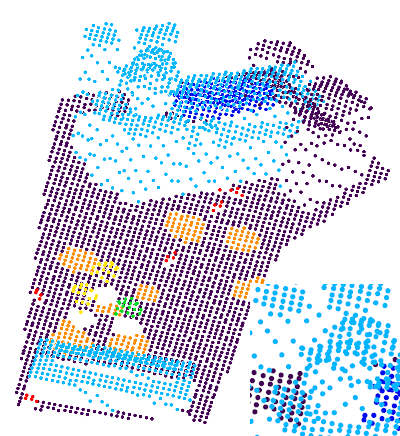} & 
        \includegraphics[width=0.3\textwidth]{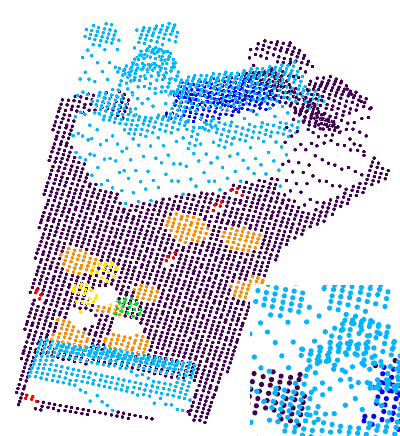}\\ 
          (\textbf{e}) OneConvFusion & (\textbf{f}) Triplet KPConv & (\textbf{g}) Encoder Fusion SiamKPConv\\
        
        \multicolumn{3}{c}{        \begin{tikzpicture}
    		\begin{axis}[
            		xmin=1,
                    xmax=2,
                    ymin=1,
                    ymax=2,
                     hide axis,
    				width=0.5\textwidth ,
    				mark=circle,
    				scatter,
    				only marks,
    				legend entries={Unchanged, New Building, Demolition, New Vegetation, Vegetation Growth, Missing Vegetation, Mobile Objects},
    				legend cell align={left},
    				legend style={draw=lightgray,at={(0,0)}, legend columns=7,/tikz/every even column/.append style={column sep=0.2cm}}]
    			\addplot[Unchanged] coordinates {(0,0)}; 
    			\addplot[NewBuild] coordinates {(0,0)};
    			\addplot[Demol] coordinates {(0,0)};
    			\addplot[VegeN] coordinates {(0,0)};
    			\addplot[VegeG] coordinates {(0,0)};
    			\addplot[VegeR] coordinates {(0,0)};
    			\addplot[MOch] coordinates {(0,0)};
    		\end{axis}
	    \end{tikzpicture}}\\
    \end{tabular}
    
    \caption[Visual change detection results of the three Siamese KPConv evolutions]{\textbf{Visual change detection results on the Urb3DCD-V2 low-density \ac{LiDAR} sub-dataset:} (\textbf{a}-\textbf{b}) two input point clouds; (\textbf{c}) ground truth (GT): simulated changes; (\textbf{d}) Siamese \ac{KPConv} results; (\textbf{e}) OneConvFusion results; (\textbf{f}) Triplet \ac{KPConv} results; and (\textbf{g}) Encoder Fusion SiamKPConv results. The region of interest specifically discussed in the text was highlighted with an ellipse.}
    \label{fig:resLid05newArchi}
\end{figure*}

\begin{figure*}
    \centering
    \footnotesize
    \begin{tabular}{cc}
        \includegraphics[ width=0.33\textwidth]{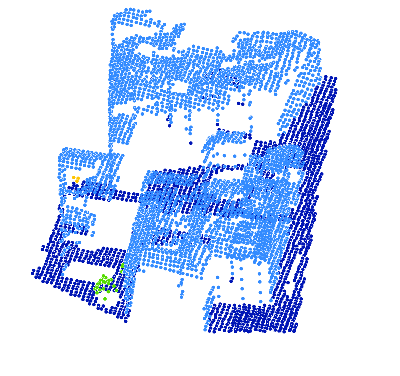} 
        & 
        \includegraphics[width=0.33\textwidth]{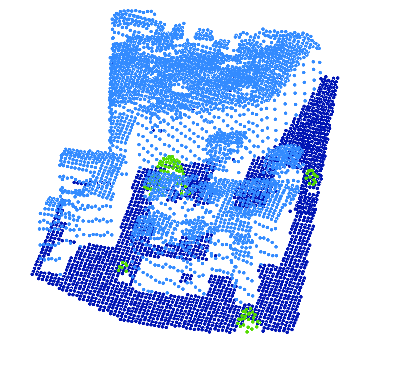} 
        \\
        \multicolumn{2}{c}{
        \begin{tikzpicture}
    		\begin{axis}[
            		xmin=1,
                    xmax=2,
                    ymin=1,
                    ymax=2,
                     hide axis,
    				width=0.3\columnwidth ,
    				mark=circle,
    				scatter,
    				only marks,
    				legend cell align={left},
    				legend entries={Ground, Building, Vegetation, Mobile Objects},
    				legend style={draw=lightgray,at={(0,0)}, legend columns=4,/tikz/every even column/.append style={column sep=0.5cm}}]
    			\addplot[Ground] coordinates {(0,0)}; 
    			\addplot[Building] coordinates {(0,0)};
    			\addplot[Vegetation] coordinates {(0,0)};
    			\addplot[MO] coordinates {(0,0)};
    		\end{axis}
	    \end{tikzpicture}} 
	    \\
        (\textbf{a}) PC 1 & (\textbf{b}) PC 2 \\
        \end{tabular}
        \begin{tabular}{cc}
        
        \includegraphics[width=0.34\textwidth]{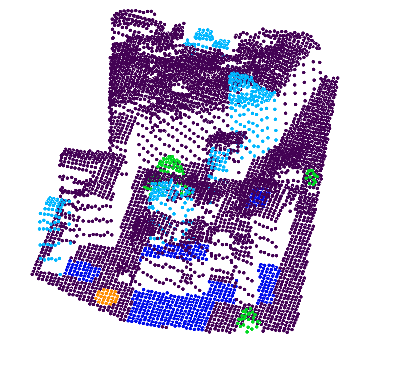} &
         \begin{tikzpicture}
    \node (image) at (0,0) {\includegraphics[width=0.34\textwidth]{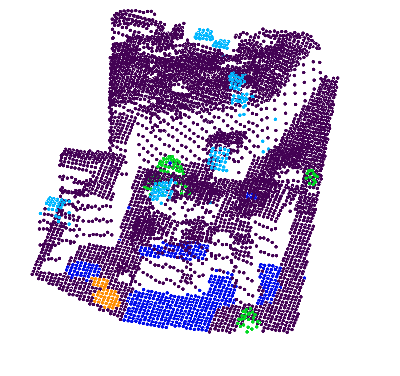}};
       \draw[red, very thick,rotate=0] (0.9,1.) ellipse (13pt and 19pt);
       \draw[red, very thick,rotate=0] (-2.2,-0.8) ellipse (11pt and 19pt);
 \end{tikzpicture}
       \\
         (\textbf{c}) GT & (\textbf{d}) Siamese KPConv \cite{degelis2023siamese} \\
        \end{tabular}
        \begin{tabular}{ccc}
         
        \includegraphics[width=0.34\textwidth]{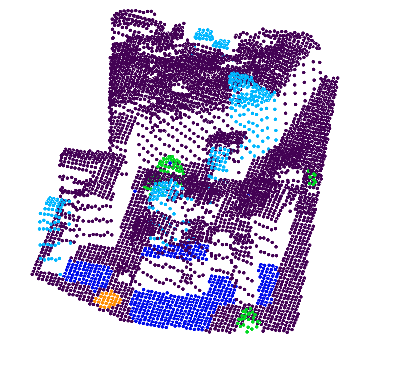}&
        \includegraphics[width=0.34\textwidth]{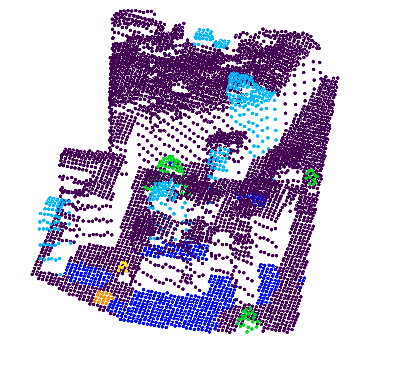} &
        \includegraphics[width=0.34\textwidth]{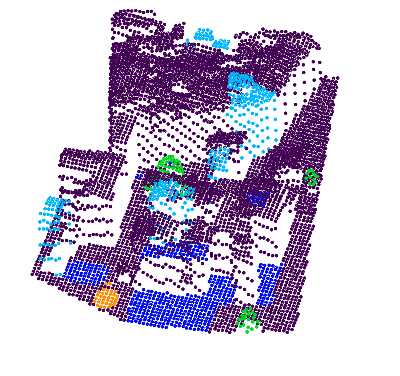}\\ 
         (\textbf{e}) OneConvFusion & (\textbf{f}) Triplet KPConv &(\textbf{g}) Encoder Fusion SiamKPConv\\
        \multicolumn{3}{c}{        \begin{tikzpicture}
    		\begin{axis}[
            		xmin=1,
                    xmax=2,
                    ymin=1,
                    ymax=2,
                     hide axis,
    				width=0.3\columnwidth ,
    				mark=circle,
    				scatter,
    				only marks,
    				legend entries={Unchanged, New Building, Demolition, New Vegetation, Vegetation Growth, Missing Vegetation, Mobile Objects},
    				legend cell align={left},
    				legend style={draw=lightgray,at={(0,0)}, legend columns=7,/tikz/every even column/.append style={column sep=0.2cm}}]
    			\addplot[Unchanged] coordinates {(0,0)}; 
    			\addplot[NewBuild] coordinates {(0,0)};
    			\addplot[Demol] coordinates {(0,0)};
    			\addplot[VegeN] coordinates {(0,0)};
    			\addplot[VegeG] coordinates {(0,0)};
    			\addplot[VegeR] coordinates {(0,0)};
    			\addplot[MOch] coordinates {(0,0)};
    		\end{axis}
	    \end{tikzpicture}}\\
    \end{tabular}
    
    \caption[Visual change detection results of the three Siamese KPConv evolutions in an area containing occlusions]{\textbf{Visual change detection results on the Urb3DCD-V2 low density \ac{LiDAR} sub-dataset in an area containing occlusions:} (\textbf{a}-\textbf{b}) two input point clouds; (\textbf{c}) ground truth (GT): simulated changes; (\textbf{d}) Siamese \ac{KPConv} results; (\textbf{e}) OneConvFusion results; (\textbf{f}) Triplet \ac{KPConv} results; and (\textbf{g}) Encoder Fusion SiamKPConv results. The regions of interest specifically discussed in the text were highlighted with ellipses.}
    \label{fig:resLid05Cas2newarchi}
\end{figure*}

Tables~\ref{tab:ResLid05Newarchi} and \ref{tab:ResLid05IouNewArchi} present the quantitative results
of the evaluation of the three architectures on the Urb3DCD-V2 dataset. Each of the three architectures outperformed the Siamese \ac{KPConv} network. The best architecture was Encoder Fusion SiamKPConv, followed by Triplet KPConv. OneConvFusion performed only slightly better with a 1.5\% of the mIoU$_{ch}$ when compared with Siamese KPConv. Looking at the per-class results (Table~\ref{tab:ResLid05IouNewArchi} and Figure~\ref{fig:influArchi}), the Encoder Fusion SiamKPConv network provided a significant improvement for all change classes. Figures~\ref{fig:resLid05newArchi} and \ref{fig:resLid05Cas2newarchi} depict the qualitative results. The three architectures provided very similar results to the ground truth. In Figure~\ref{fig:resLid05Cas2newarchi}, each of the three Siamese \ac{KPConv} evolutions showed results that were more accurate than those in Siamese \ac{KPConv} in the new building facades. These facades were particularly hard to correctly detect because the neighboring facade was not visible in the first \ac{PC} (Figure~\ref{fig:resLid05Cas2newarchi}a). 
Therefore, identifying the new facade in the class `new building' while the neighboring facades were unchanged was not obvious.
In this situation, the network should understand that the facade may be new because the roof is new. In the same manner, a roof that remains unchanged should also have an unchanged facade. Another difference with the Siamese \ac{KPConv} results can be found in Figure~\ref{fig:resLid05newArchi} (zoomed out portions), where a part of the church roof is identified as new vegetation for Siamese \ac{KPConv} only, not for the other architectures. The misclassification was probably due to the dome roof shape that looked like a tree in the simulated data. Even if the tree models were not totally spherical (i.e., the Arbraro software \cite{diestel2003arbaro} was used to obtain OBJ models of trees, see \cite{degelis2023siamese}), the \ac{LiDAR} simulation on these models rendered a quite spherical object with only a few points inside the foliage of the tree, unlike the real \ac{LiDAR} acquisition. Therefore, aside from the shape, the main factor for distinguishing between vegetation and the dome is that trees are generally on the ground. These examples highlight the fact that the network should be able to understand the \ac{PC} at multiple scales and predict changes with regard to the surrounding objects.

\begin{table}
    \centering
      \caption[Results of the three Siamese \ac{KPConv} evolutions]{\textbf{Results in \% of the three Siamese \ac{KPConv} evolutions on the Urb3DCD-V2 low density \ac{LiDAR} dataset.}}
    \label{tab:ResLid05Newarchi}
    \footnotesize
    \begin{tabular}{c|cc}
    \hline
         Method & mAcc (\%)& mIoU$_{ch}$ (\%)\\
    
        \hline
             Siamese KPConv \cite{degelis2023siamese}  & 91.21 $\pm$ 0.68 & 80.12 $\pm$ 0.02 \\
             Siamese KPConv  (+10 input features)& 93.65 $\pm$ 0.16 & 84.82 $\pm$ 0.58 \\
        OneConvFusion & 92.62 $\pm$ 1.10 & 81.74 $\pm$ 1.45\\
        Triplet KPConv  & 92.94 $\pm$ 0.53 & 84.08 $\pm$ 1.20 \\
           Encoder Fusion SiamKPConv  & \textbf{94.23} $\pm$ 0.88 & \textbf{85.19} $\pm$ 0.24\\
        \hline
    \end{tabular}
  
\end{table}

\begin{table*}
    
   \centering
    \caption[Per-class IoU scores of the three Siamese KPConv evolutions]{\textbf{Per-class \ac{IoU} scores of the three Siamese \ac{KPConv} evolutions on the Urb3DCD-V2 low density \ac{LiDAR} dataset.} The results were given in \%. Veg.: vegetation; input feat.: input features; and SKPConv: Siamese KPConv. }
     
     \footnotesize
        \begin{tabular}{c|ccccccc}
        \hline
         \multirow{2}{*}{Method} & \multicolumn{7}{c}{Per class IoU (\%)}\\
         & Unchanged & New building & Demolition & New veg. & Veg. growth& Missing veg.& Mobile Object\\
        \hline
        Siamese KPConv \cite{degelis2023siamese}& 95.82 $\pm$ 0.48 & 86.67 $\pm$ 0.47 & 78.66 $\pm$ 0.47 & 93.16 $\pm$ 0.27 & 65.18 $\pm$ 1.37 & 65.46 $\pm$ 0.93 & 91.55 $\pm$ 0.60\\
        SKPConv ($+ 10$ input feat.) & \textbf{97.55} $\pm$ 0.11 & 95.17 $\pm$ 0.21 & \textbf{84.25} $\pm$ 0.59 & 95.23 $\pm$ 0.21 & 66.02 $\pm$ 1.33 & \textbf{74.88} $\pm$ 1.03 & 93.38 $\pm$ 0.74\\
        OneConvFusion & 96.95 $\pm$ 0.34 & 96.06 $\pm$ 0.27 & 79.63 $\pm$ 1.48 & 95.53 $\pm$ 0.77 & 61.12 $\pm$ 2.13 & 65.79 $\pm$ 2.61 & 92.89 $\pm$ 1.95\\
        Triplet KPConv & 97.41 $\pm$ 0.24 & 95.73 $\pm$ 0.67 & 81.71 $\pm$ 1.47 & 96.24 $\pm$ 0.37 & 64.85 $\pm$ 1.46 & 73.02 $\pm$ 1.18 & 92.90 $\pm$ 2.47\\
        Encoder Fusion SKPConv & 97.47 $\pm$ 0.04 & \textbf{96.68} $\pm$ 0.30 & 82.29 $\pm$ 0.16 & \textbf{96.52} $\pm$ 0.03 & \textbf{67.76} $\pm$ 1.51 & 73.50 $\pm$ 0.81 & \textbf{94.37} $\pm$ 0.54\\
            
             \hline
        \end{tabular}

    \label{tab:ResLid05IouNewArchi}
    \end{table*}
    
Table~\ref{tab:ResLid05NewarchiAHNCD} presents the quantitative results for the experiments on the real data. All proposed architectures and exploitation of the hand-crafted features enabled us to improve the state-of-the-art Siamese KPConv results. Encoder Fusion SiamKPConv and OneConvFusion showed the largest improvements of up to approximately 5\% of the mIoU$_{ch}$. In contrast to the results on the Urb3DCD-V2 results, OneConvFusion obtained results comparable with those of Encoder Fusion SiamKPConv, albeit with a larger standard deviation. Given that the training set for \ac{AHN-CD} included numerous labeling errors and considering the standard deviations of OneConvFusion and Encoder Fusion SiamKPConv, these two approaches were similar in terms of performance. Although OneConvFusion achieved only a minor enhancement in Urb3DCD-V2, it produced results that yielded a significant improvement on the \ac{AHN-CD} dataset similar to Encoder Fusion SiamKPConv. OneConvFusion is a network with fewer parameters compared to other methods (Table~\ref{tab:nbParams}). This probably led to a better generalization of the training data and ended up in superior results despite the numerous errors present in the training database.

\begin{table}[htbp]
    \centering
      \caption[Results of the three Siamese \ac{KPConv} evolutions]{\textbf{Results in \% of the three Siamese \ac{KPConv} evolutions on the manually cleaned \ac{AHN-CD} dataset.}}
    \label{tab:ResLid05NewarchiAHNCD}
    \footnotesize
    \begin{tabular}{c|cc}
    \hline
         Method & mAcc (\%)& mIoU$_{ch}$ (\%)\\
    
        \hline
             Siamese KPConv \cite{degelis2023siamese}  & 85.65 $\pm$ 1.55 & 70.65 $\pm$ 2.05 \\
             Siamese KPConv  (+10 input features)& 88.47 $\pm$ 1.09 & 73.29 $\pm$ 1.32 \\
        OneConvFusion & 90.03 $\pm$ 0.38 & \textbf{75.62} $\pm$ 1.04\\
        Triplet KPConv  & 88.25 $\pm$ 0.23 & 72.37 $\pm$ 0.55 \\
           Encoder Fusion SiamKPConv  & \textbf{90.26} $\pm$ 0.22 & 75.00 $\pm$ 0.74\\
        \hline
    \end{tabular}
  
\end{table}

\section{Discussion}\label{discu}

\subsection*{Number of parameters}
While it was important to study the predicted change detection results, it is also relevant to check the number of trainable parameters in each architecture. Table~\ref{tab:nbParams} shows that Triplet KPConv and Encoder Fusion SiamKPConv required approximately two times more trainable parameters, whereas OneConvFusion needed exactly the same number of parameters as Siamese KPConv. Adding input features only slightly increased the number of parameters. This must be considered in case of limited training capabilities. Directly guiding the network to change detection by introducing hand-crafted features seems interesting in preventing a drastic increase of trainable parameters.  
\begin{table}
    \centering
      \caption{\textbf{Number of parameters in each presented architecture compared to those of the original Siamese KPConv network.}}
    \label{tab:nbParams}
    \footnotesize
    \begin{tabular}{c|c}
    \hline
        Method & \multicolumn{1}{c}{Number of parameters}\\
        \hline
             Siamese KPConv \cite{degelis2023siamese}  & 18,441,152 \\ 
             Siamese KPConv  (+10 input features)& 18,457,152 \\ 
        OneConvFusion & 18,441,152 \\ 
        Triplet KPConv  & 39,753,536 \\ 
           Encoder Fusion SiamKPConv  & 39,841,152 \\ 
        \hline
    \end{tabular}
  
\end{table}

\subsection*{Importance of learning change information}

An immediate observation from our experiments was that adding hand-crafted features related to both input data as the input to the Siamese \ac{KPConv} network did not bring any significant change on the results (Table \ref{tab:ResLid05Newarchi}, third line --Addition of features-- compared to the first one). 

Looking at the results presented in \cite{degelis2023siamese}, the Siamese \ac{KPConv} architecture was able to detect the change on its own; however, it seems that giving a hand-crafted feature related to the change specifically as an input helps the network focus on the change with a significant improvement (Table \ref{tab:ResLid05Newarchi}, fourth  line --Addition of a change feature-- compared to the first one). 

This finding underlines the fact that encoding change information is important. On this basis, we confirm that the proposed evolutions of Siamese KPConv depict the relevance of applying convolution on the nearest-point feature difference at multiple scales to obtain change-related features (Table \ref{tab:ResLid05IouNewArchi} three last lines) not by introducing change features, but by encoding them directly. The somewhat worse results of the OneConvFusion network, specifically on \ac{Urb3DCD}, exhibit the importance of keeping multi-scale mono-date features in the architecture. Note that this is in line with the deep learning for the change detection literature in \ac{2D}\cite{daudt2018fully, chen2019deepSiam, zhang2020feature, zheng2023global}. The fact that Encoder Fusion SiamKPConv provided better results compared to the Triplet network showed that combining both mono-date semantic and change features as input to the convolutional layers enables the extraction of useful discriminative features for the change segmentation task. Note that both Triplet \ac{KPConv} and Encoder Fusion SiamKPConv obatined results close to (or even outperformed) the Siamese \ac{KPConv} network with the 10 hand-crafted features. 

Finally, we added hand-crafted features as the Encoder Fusion SiamKPConv network input. The results were only very slightly improved at less than 1\% of the mIoU$_{ch}$. In other words, an architecture more specifically designed for change detection can extract discriminative features on its own. This is especially true for the change-related feature of $Stability$, which is no longer required in the Encoder Fusion SiamKPConv architecture.

\section{Conclusion}\label{sec:ccl}
In this work, we proposed the enhancement of change detection in raw \ac{3D} \acp{PC} by using deep networks. We suggested the introduction of the change information early in the network to better detect and categorize the changes in 3D \acp{PC}. The first proposition for enhancing the existing method was to provide some hand-crafted features as input along with 3D point coordinates. We demonstrated that  the addition of a single change-related feature input to the existing Siamese KPConv method yields an enhancement of approximately 3.70\% of the mean of the IoU over change classes. We also proposed three new architectures for change segmentation based on raw 3D \acp{PC} that encoded the change information conversely to the current state-of-the-art, which only incorporated change information in the decoder step. All three architectures out-performed the current SOTA methods by up to 5.07\% of the mean of the IoU over classes of change. In conclusion, the importance of encoding change information was confirmed in this investigation.

As for future work, we mainly see two axes of improvement. First one could imagine introducing attention mechanism for multi-scale fusion of both change-related and mono-date features, as was already successfully shown in recent studies in \ac{2D} image change detection \cite{fang2021snunet,chen2022msf,YIN2023103206}. Investigating transformers \cite{guo2021pct} is an interesting perspective for the improvement of our method. However, their scaling to large remote sensing \acp{PC} is often not obvious. Finally, these architectures need to be trained on large amounts of data to obtain convincing results. Considering the difficulty of \ac{PC} annotation, the second axis of improvement is to focus on semi-supervised methods for the \ac{3D} change detection.

\begin{acronym}
    \acro{1D}{one-dimensional}
    \acro{2D}{two-dimensional}
    \acro{3D}{three-dimensional}
    \acro{AHN-CD}{Actueel Hoogtebestand Nederland Change Detection}
    \acro{ALS}{Aerial Laser Scanning}
    \acro{DSM}{Digital Surface Model}
    \acro{DTM}{Digital Terrain Model}
    \acro{FCN}{Fully Convolutionnal Network}
    \acro{GPU}{graphics processing unit}
    \acro{IoU}{Intersection over Union}
    \acro{kNN}{k-Nearest Neighbors}
    \acro{KPConv}{Kernel Point Convolution}
    \acro{KP-FCNN}{Kernel Point -- Fully Convolutional Neural Network}
    \acro{LiDAR}{Light Detection And Ranging}
    \acro{mAcc}{mean accuracy}
    \acro{mIoU}{mean of \ac{IoU}}
    \acro{MLS}{Mobile Laser Scanning}
    \acro{NLL}{negative log-likelihood}
    \acro{PC}{Point Cloud}
    \acro{PCA}{Principal Component Analysis}
    \acro{PDAL}{Point Data Abstraction Library}
    \acro{RF}{Random Forest}
    \acro{SGD}{Stochastic Gradient Descent}
    \acro{Urb3DCD}{Urban 3D Change Detection}
    \acro{Urb3DCD-Cls}{Urban 3D Change Detection Classification}
\end{acronym}

\section*{Acknowledgment}

This research was funded by Magellium and the CNES, Toulouse, with access to the HPC resources of IDRIS under the allocation 2022-AD011011754R2 made by GENCI.

\ifCLASSOPTIONcaptionsoff
  \newpage
\fi



%



\bibliography{biblio}
\bibliographystyle{IEEEtran}
%








\end{document}